%% file: main.tex
\documentclass[runningheads]{llncs}
\usepackage[T1]{fontenc}
\usepackage{graphicx}
\usepackage[hidelinks]{hyperref}
\usepackage{color}

\usepackage{amsmath}
\usepackage{amssymb}

\DeclareMathOperator*{\argmin}{arg\,min}
\usepackage[ruled,vlined,algo2e]{algorithm2e}
\SetAlFnt{\small}
\usepackage{booktabs}
\usepackage{siunitx}
\usepackage{multirow}
\newcommand*{\thead}[1]{\multicolumn{1}{c}{\bfseries\begin{tabular}{@{}c@{}}#1\end{tabular}}}
\usepackage{subcaption}

\begin{document}
\title{Elastic Product Quantization for Time Series}
%
%
\author{Pieter Robberechts\inst{1} \orcidID{0000-0002-3734-0047} \and
Wannes Meert\inst{1}\orcidID{0000-0001-9560-3872} \and
Jesse Davis\inst{1}\orcidID{0000-0002-3748-9263}}
\authorrunning{P. Robberechts et al.}
%
\institute{KU Leuven, Dept. of Computer Science; Leuven.AI, B-3000 Leuven, Belgium
\email{\{pieter.robberechts,wannes.meert,jesse.davis\}@kuleuven.be}}
\maketitle              
\input{chapters/00-abstract}
\input{chapters/01-introduction}
\input{chapters/02-background}
\input{chapters/03-method}
\input{chapters/04-experiments}
\input{chapters/07-conclusion}

\section*{Acknowledgements}
This work was partially supported by iBOF/21/075, the KU Leuven Research Fund (C14/17/070), VLAIO ICON-AI Conscious, and the Flemish Government under the “Onderzoeksprogramma Artificiële Intelligentie (AI) Vlaanderen” program.
%
%
%
\bibliographystyle{splncs04}
\bibliography{references}
\end{document}

%% file: chapters/00-abstract.tex
\begin{abstract}
Analyzing numerous or long time series is difficult in practice due to the high storage costs and computational requirements.
Therefore, techniques have been proposed to generate compact similarity-preserving representations of time series, enabling real-time similarity search on large in-memory data collections. 
However, the existing techniques are not ideally suited for assessing similarity when sequences are locally out of phase.
In this paper, we propose the use of product quantization for efficient similarity-based comparison of time series under time warping. The idea is to first compress the data by partitioning the time series into equal length sub-sequences which are represented by a short code.
The distance between two time series can then be efficiently approximated by pre-computed elastic distances between their codes. 
The partitioning into sub-sequences forces unwanted alignments, which we address with a pre-alignment step using the maximal overlap discrete wavelet transform (MODWT). 
To demonstrate the efficiency and accuracy of our method, we perform an extensive experimental evaluation on benchmark datasets in nearest neighbors classification and clustering applications.
Overall, the proposed solution emerges as a highly efficient (both in terms of memory usage and computation time)
replacement for elastic measures in time series applications.
\end{abstract}

%% file: chapters/01-introduction.tex
\section{Introduction}


Data mining applications on large time series collections are constrained by the computational cost of similarity comparisons between pairs of series and memory constraints on the processing device. The general approach to overcome these constraints is to first apply a transformation that produces a compact representation of the time series that retains it's main characteristics. Many techniques have been proposed to generate such representations for time series analysis, including techniques based on Discrete Fourier Transform (DFT)~\cite{faloutsos1994:DFT}, Discrete Wavelet Transform (DWT)~\cite{chang99:DWT}, Singular Value Decomposition (SVD)~\cite{chan03:SVD}, and segmentation~\cite{keogh2001PAA}.

Most of these techniques are based on the Euclidean distance as the metric for similarity. However, there are some cases where the Euclidean distance,
and lock-step measures in general, may not be entirely adequate for estimating similarity~\cite{paparrizos2020}. The reason is that the Euclidean distance is sensitive to distortions along the time axis. To avoid this problem, similarity models should allow some elastic shifting of the time dimension to detect similar shapes that are not locally aligned. This is resolved by elastic measures such as Dynamic Time Warping (DTW)~\cite{sakoe1978}. The ability to accommodate temporal aberrations comes at a cost however, as the standard dynamic programming approach for computing the DTW measure has a quadratic computational complexity.

To accommodate efficient {\it approximate} nearest neighbor (NN) search in large datasets under time warping, Zhang et al.~\cite{zhang2021} proposed to adapt the technique of product quantization (PQ)~\cite{jegou2010pq}. In the standard feature-vector case, PQ is extremely performant for approximate NN search based on Euclidean distance. Its core idea is to (1) partition the vectors into disjoint subspaces, (2) cluster each subspace independently to learn a codebook of centroids, (3) re-represent each vector by a short code composed of its indices in the codebook, and (4) efficiently conduct the search over these codes using look-up tables. This has the dual benefits of greatly shrinking the memory footprint of the training set while simultaneously dramatically reducing the number of computations needed.

Unfortunately, naively combining the conventional PQ with DTW results in both missed and unwanted alignments. First, PQ segments a time series and the optimal alignment can cross segments. Second, applying DTW on a segment forces an unwanted alignment at the beginning and end of a segment. 
Zhang et al.~\cite{zhang2021} got round these challenges with a {\it filter-and-refine} post-processing step that calculates the exact DTW distances between the best candidate time series to filter out the erroneous alignments. However, this {\it increases} the memory footprint of NN-DTW (as both the original training set and a compact representation have to be retained) and does not solve the problem of false dismissals due to missed alignments. Instead, we propose a pre-alignment step using the maximal overlap discrete wavelet transform (MODWT). The resulting method is an approximate DTW method that is fast in its own right while it can still benefit from previous advances in speeding-up DTW, such as constraint bands and pruning strategies.

 
The contributions of this paper are as follows:
(1)~We bridge the gap between PQ and DTW, introducing a pre-alignment step to minimize the effect of segmentation that is part of PQ on DTW;
(2)~We show how our method is compatible with and speeds up tasks such as nearest neighbours and clustering;
(3)~We demonstrate empirically the utility of our approach by comparing it to the most common distance measures on the ubiquitous UCR benchmarks.

%% file: chapters/02-background.tex
\section{Background}
A large body of literature is available on dynamic time warping and product quantization. In this section, we restrict our presentation to the notations and concepts used in the rest of the paper.

\subsection{Dynamic Time Warping}

Dynamic Time Warping (DTW) computes the distance between two time series $A$ and $B$ after optimal alignment \cite{sakoe1978}. The alignment is computed using (1) dynamic programming with
\begin{align*}
\textrm{dtw\_dist}[i,j] = (A_i - B_j)^2 + \min\begin{cases}
\textrm{dtw\_dist}[i-1,j-1]\\
\textrm{dtw\_dist}[i,j-1]\\
\textrm{dtw\_dist}[i-1,j]
\end{cases}
\end{align*}
where $i$ and $j$ are indices for $A$ and $B$, and (2) the minimum cost path in the matrix $\textrm{dtw\_dist}[i,j]$. The value $DTW(A,B) = \textrm{dtw\_dist}[\textrm{length}(A),\ \textrm{length}(B)]$ is the distance between series $A$ and $B$.
DTW is particularly useful for comparing the shapes of time series, as it compensates for subtle variations such as shifts, compression and expansion.

The standard dynamic programming approach for computing the DTW measure has a quadratic computational complexity. Four common approaches exist to enable scaling to large datasets:
constraint bands or warping windows~\cite{sakoe1978}, lower-bound pruning~\cite{kim2001lbkim,keogh2005lbkeogh,shen2018lbnew,tan2019elastic,lemire2009lbimproved,rakthanmanon2012lbcascade}, pruning warping alignments~\cite{silva2016pruneddtw}, and DTW approximations~\cite{salvador2007fastdtw,spiegel2014ltw}. While proposed methods across these four approaches reduce the theoretical complexity of DTW down from quadratic, they incur other costs. Constraining bands decreases accuracy, especially in domains with large distortions. Lower-bound pruning is increasingly efficient for larger datasets, but cannot be applied for tasks such as clustering since it requires the complete pairwise distance matrix. Pruning warping alignments is a valuable addition but has limited impact when many suitable warping paths exist. Approximate methods introduce an additional complexity that requires more memory and loses the computational simplicity of the original DTW algorithm making it often slow in practice~\cite{wu2020slowdtw}. In contrast, our proposed approach is fast in its own right while maintaining compatibility with the aforementioned techniques.

\subsection{Product Quantization}
Product Quantization (PQ)~\cite{jegou2010pq} is a well-known approach for approximate nearest neighbors search for standard feature-vector data using the Euclidean distance. It confers two big advantages. One, it can dramatically compress the size of the training set, which enables storing large datasets in main memory. Two, it enables quickly computing the approximate Euclidean distance between a test example and each training example. 

PQ compresses the training data by partitioning each feature vector used to describe a training example into $M$ equal sized groups, termed {\it subspaces}. It then learns a {\it codebook} for each subspace. Typically, this is done by running k-means clustering on each subspace which only considers the features assigned to that subspace. Then the values of all features in the subspace are replaced by a single $v$-bit code representing the id of the cluster centroid $c_k$ that the example is assigned to in the current subspace. Hence, each example is re-represented by $M$ $v$-bit code words. This mapping is termed the {\it quantizer}. 

At test time, finding a test example's nearest neighbor using the squared Euclidean distance can be done efficiently by using table look-ups and addition. For a test example, a look-up table is constructed for each subspace. This table stores the squared Euclidean distance to each of the $K$ 
cluster centroids in that subspace. Then the approximate distance to each training example is computed using these look-up tables and the nearest example is returned. 

%% file: chapters/03-method.tex
\section{Approximate Dynamic Time Warping with Product Quantization}
This section introduces the DTW with Product Quantization (PQDTW) approach~\cite{zhang2021}. First, we discuss how each component of the original PQ method can be adapted to a DTW context. This encompasses learning the codebook, encoding the data and computing approximate distances between codes. Then, we extend the base method to compensate for the alignment loss caused by partitioning the time series. Finally, we explain how PQDTW can be used in NN search and clustering.

\subsection{Training phase}

The training phase 
comprises the learning of a codebook using DTW Barycenter Averaging (DBA) k-means~\cite{petitjean2011}. Let us consider a training set of time series $X=\left[x_1, x_2, \ldots, x_N\right] \in \mathbb{R}^{N \times D}$ (i.e., $N$ time series of length $D$). Each time series in the dataset is first partitioned in $M$ sub-sequences, each of length $D/M$. Subsequently, a sub-codebook for each $m \in \{1,\ldots,M\}$ is computed: $C^m = \{c^m_k\}^K_{k=1}$, with centroids $c^m_{k} \in \mathbb{R}^{D/M}$. Each $C_m$ is obtained by running the DBA k-means clustering over the $m$\textsuperscript{th} part of the training sequences, where $K$ is the number of clusters. The $K \times M$ centroids obtained by k-means represent the most commonly occurring patterns in the training time series' subspaces.

Two other pre-processing steps can be performed during the training phase to speed-up the encoding of time series and computing the symmetric distances between codes: the construction of the Keogh envelopes~\cite{keogh2005lbkeogh} of all centroids and the computation of a distance look-up table with the DTW distance between each pair of centroids in a subspace. These steps are explained in the next sections.

The training step has to be performed once to optimize it for a specific type of data, but can be reused to speed up computations on future examples from the same domain.

\subsection{Encoding time series}
\label{sec:encoding}
Using the codebook, we can represent any time series as a short code. The idea is to (1) partition a time series into sub-sequences, (2) independently encode each sub-sequence to an identifier, and (3) re-represent the time series as a concatenation of the identifiers. A given time series $x \in \mathbb{R}^{D}$ is therefore mapped as:
\begin{align*}
x \rightarrow \left[q_1(x_1, \ldots x_{D/M}), \ldots, q_{M}(x_{D-D/M+1},\ldots,x_{D})\right],
\end{align*}
\noindent where $q_m: \mathbb{R}^{D/M} \rightarrow \{1,\ldots, K\}$ is the quantizer associated with the $m$\textsuperscript{th} subspace that maps a sub-sequence $$x^m = \left(x_{m\times(D/M))}, \ldots x_{(m+1)\times(D/M)}\right)$$ to the identifier of the nearest centroid $c^m_k$ in the codebook. Formally, this search is defined as:
\begin{align*}
    q_m(x^m) = \argmin_{k \in \{1, \ldots, K\}}\mathrm{DTW}\left(x^m,\ c^m_k\right).
\end{align*}

Practically, this search is performed by linearly comparing a $D/M$-dimensional sub-sequence to $K$ centroids (i.e., a NN-DTW query), which has a computational complexity of $O(K\times(D/M)^2)$ with standard DTW. Since the sub-sequences are quantized separately using $M$ distinct quantizers, the overall computational complexity of encoding a time series is $O(K\times D^2/M)$. 

The quadratic complexity of the dynamic programming approach to DTW makes the NN-DTW queries required to encode a time series highly computationally demanding. In regular NN-DTW, cheap-to-compute lower bounds are a key strategy to combat this by pruning the expensive DTW computations of unpromising nearest neighbour candidates \cite{rakthanmanon2012lbcascade}. This involves computing an enclosing envelope around the query (i.e., the sub-sequence to be encoded) which is reused to compute the actual lower bound between the query and each test time series. This would be inefficient in our use case, since it would require the construction of the envelope every time the PQ is used to encode a time series. Therefore, we reverse the query/data role in the lower bound search~\cite{rakthanmanon2012lbcascade}. This enables computing the envelopes only once around the codebook, which can be done during the training phase.

Many DTW lower bounds have been proposed, including LB Kim~\cite{kim2001lbkim}, LB Keogh~\cite{keogh2005lbkeogh}, LB Improved~\cite{lemire2009lbimproved}, LB New~\cite{shen2018lbnew} and LB Enhanced~\cite{tan2019elastic}; as well as cascading lower bounds that start with a looser (and computationally cheap) one and progress towards tighter lower bounds~\cite{rakthanmanon2012lbcascade}. In the experimental evaluation of this paper, we use a cascading lower bound of LB Kim and the reversed LB Keogh, which provides an effective trade-off between speed and tightness for small window sizes (i.e, as we obtain after partitioning the time series)~\cite{tan2019elastic}. Nevertheless, other bounds might be more effective depending on the time series' properties. Given the pre-computed upper and lower envelopes at training time, the cost of computing these bounds is only $O(D/M)$.

While this encoding is the most costly part of the PQDTW algorithm, it can be executed offline in many applications. For example, in NN search, the dataset can be encoded during the training phase and the costs can be amortized over multiple subsequent queries.


\subsection{Computing distances between time series}
Consider two time series $x$ and $y \in \mathbb{R}^D$ and a trained product quantizer $q$. The original PQ paper~\cite{jegou2010pq} proposes two methods to estimate the distance between these two time series: symmetric and asymmetric distance computation.


\paragraph{Symmetric distance} This method computes the distance between the PQ-codes of $x$ and $y$. Therefore, both time series are first encoded by the product quantizer. Secondly, a distance score is computed by fetching the centroid distances from $q$'s pre-computed distance table. This means that the  distances between the centroids $c_i^m$ and $c_j^m$ of $x$ and $y$ in each subspace $m$ need to be aggregated into one distance $d$ as: 
\begin{align*}
    \hat{d}(x, y) = d(q(x),\ q(y)) = \sqrt{\sum_{m=1}^{M} d(c_i^m,\ c_j^m)^2}.
\end{align*}
The distances $d(c_i^m,\ c_j^m)^2$ between each pair of centroids in a subspace are pre-computed during the training phase and stored in a $M$-by-$K$-by-$K$ look-up table.  Hence, symmetric distance computation is very efficient, taking only $O(M)$ table look-ups and additions.

\paragraph{Asymmetric distance} This method encodes only one of the two time series and estimates the distance between the PQ code of $x$ and the original series $y$ as
\begin{align*}
    \hat{d}(x, y) = d(q(x),\ y) = \sqrt{\sum_{m=1}^{M} d(c_i^m,\ y^m)^2}.
\end{align*}
The distances $d(c_i^m,\ y^m)^2$ between the $M$ centroids of $x$ and the $M$ subspaces of $y$ have to be computed on-the-fly. Hence, this method is inefficient to compute the distances between a single pair of time series. However, when computing the distances between a query time series $y$ and a database with many time series $X = \{x_n\})_{n=1}^N$, it becomes efficient to first construct a distance look-up table for each pair $(y^m,\ c_i^m)$ with $m \in \{1,\ldots,M\}$ and $i \in \{1,\ldots,K\}$. The computation of this look-up table takes $O(D\times K)$ DTW computations, and is performed just once per query. Subsequently, the distance computation itself takes only $O(M)$ table look-ups and additions.

Whether symmetric or asymmetric is the most appropriate distance measure depends on the application, the amount of data and the required accuracy.


\subsection{Memory cost}
Due to the conversion of time series to short codes, product quantization allows large time series collections to be processed in memory. Since these codes consist of $M$ integers ranging from $1$ to $K$, the parameters $M$ and $K$ control the main memory cost of our approach. Typically, $K$ is set as 256, such that each code can be represented by $8M$ bits (i.e., each integer in the code is represented by 8 bits). If a $D$-dimensional time series is represented in single-precision floating-point format (i.e., $32\times D$ bits), a PQ-code with $K=256$ compresses the original series by a factor $32D/8M = 4D/M$. For example, time series of length 140 can be represented $80\times$ more efficiently by PQ-codes with 7 subspaces. Larger values for $M$ lead to faster performance and a higher memory cost, but the effect on representation error is domain-dependent.

In addition to the data memory cost, our method requires a small amount of additional memory for storing the codebook ($32\times D\times K$ bits), pre-computed distance look-up table ($32\times K^2\times M$ bits) and Keogh envelopes ($2\times 32\times D\times K$ bits). In total, this corresponds to $32\times K \times (3\times D+K\times M)$, which is negligible in relation to the data memory cost. For our previous example, with $D = 140$, $K=256$ and $M=7$, the total cost is limited to $2.3$MB.

\subsection{Pre-alignment of subspaces}
\label{sec:shape_segmented_subspaces}

\begin{figure}[h]
  \centering
  \includegraphics[width=.6\linewidth]{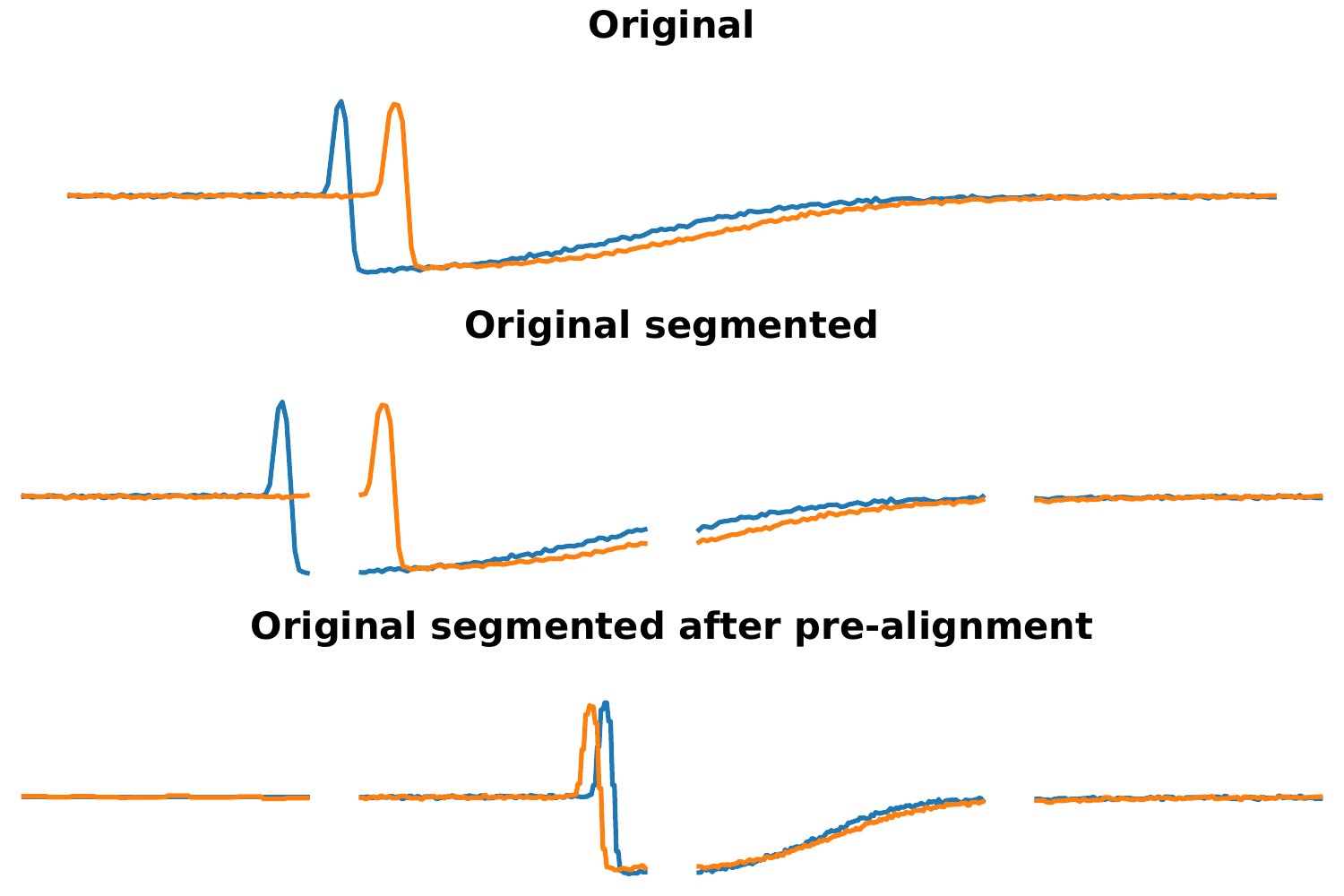}
  \caption{Segmentation of two time series of the Trace~\cite{UCRArchive2018} data set (top) in four subspaces using fixed-length subspaces (middle) and our own MODWT-based method (bottom).}
  \label{fig:trace_segmentation}
\end{figure}

When partitioning time-series into equal length sub-sequences, the endpoints of these sub-sequences might not be aligned well.
This problem is illustrated in Figure~\ref{fig:trace_segmentation}. The middle row of the plot shows the subspaces obtained by dividing two similar time series of the Trace dataset~\cite{UCRArchive2018} in four equal partitions. Notice that the distinctive peak near the first split point falls in different subspaces for both series. Because it is not possible to warp across segments, the effect is that the location of the split point will tend to contribute disproportionately to the estimated similarity, resulting in a higher approximate distance. In this section we introduce a pre-alignment step to deal with this problem.

\begin{figure}[h]
  \centering
  \includegraphics[width=\linewidth]{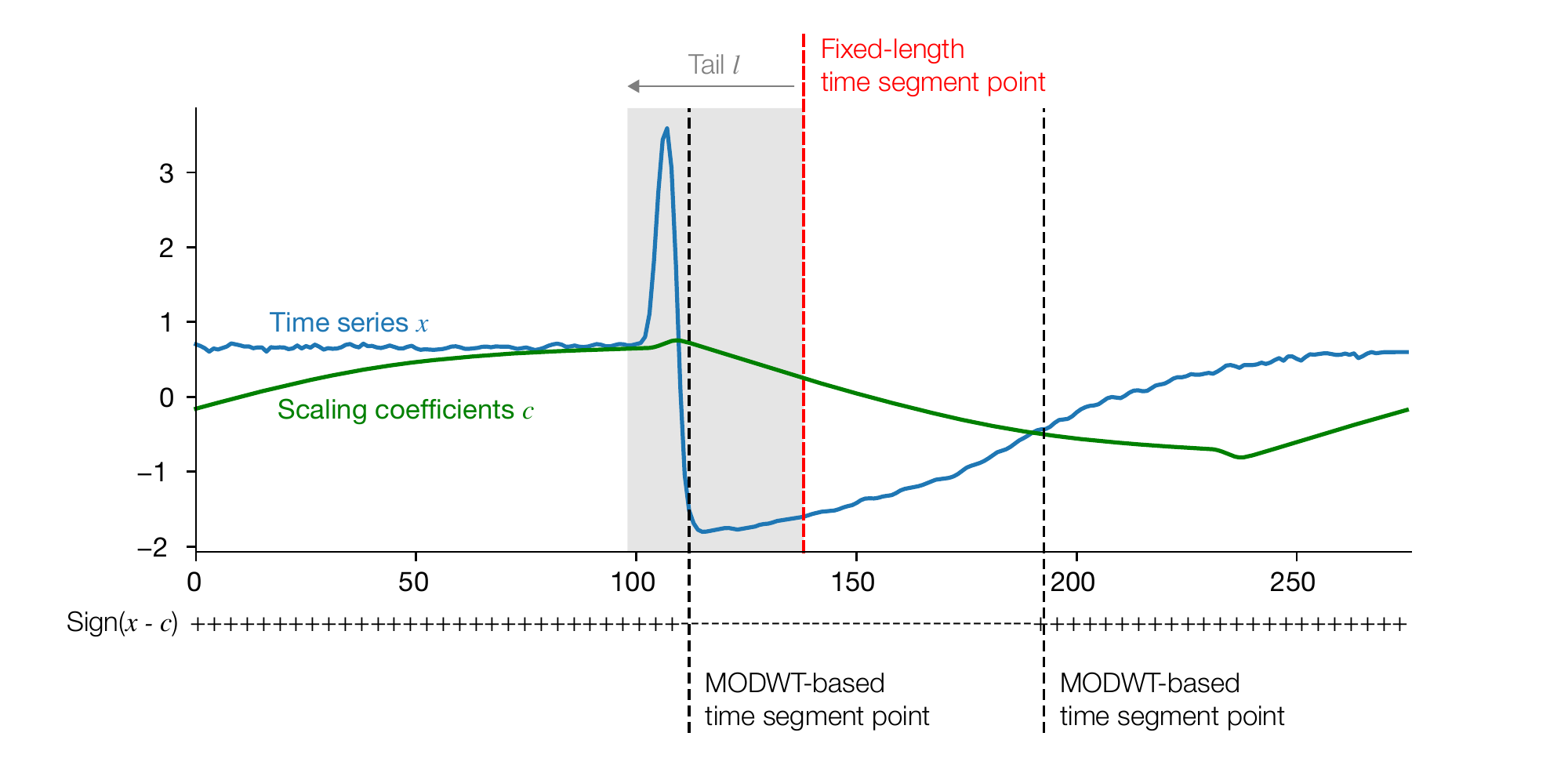}
  \caption{Illustration of the segmentation procedure based on the pre-alignment of time series. MODWT-based time segment points are extracted as the points at which the signs of the differences between the time series data and scale coefficients change. If the tail of one of the fixed length segment points contains one of these MODWT-based segment points, the MODWT-based point is used instead.}
 \label{fig:trace_segmentation_ex}
\end{figure}

The idea is to identify local structures in the time series data and segment the time series at the boundaries of these structures. For this, we use the Maximal Overlap Discrete Wavelet Transform (MODWT), as proposed by Hong et al.~\cite{hong2020SSDTW}.
Via convolution of a raw time series $x \in \mathbb{R}^D$ and the basis functions (Haar wavelet) of the MODWT, we obtain the scale coefficients $c_{j,i}$, where $j$ is the level of the decomposition $\in \{1,\ldots, J\}$ and $i \in \{1,\ldots,D\}$. These coefficients are proportional to the mean of the raw time series data. The scale coefficients of the MODWT have a length $D$ that is the same as that of the raw time series. Next, time segment points are extracted as the points at which the signs of the differences between the time series data and scale coefficients change, as shown in Figure~\ref{fig:trace_segmentation_ex}. Since the complexity of MODWT is only $O(J\times D)$, this segmentation step does not increase our method's overall complexity.

Since one time series archive typically contains series with different patterns, partitioning the time series based on local structures implies that the number of subspaces can vary between sequences. If this happens, the distances between two such series cannot be approximated by the sum of the distances between their subspaces. Therefore, a degree of fuzziness is needed to cut the time series such that each series is split in the same number of segments of similar length while avoiding splits of distinctive local structures. This is achieved by specifying a tail, $t$, measured backwards from each original fixed-length split point $l$, within which the cut should be applied; thus the cut will fall between $l-t$ and $l$. If the MODWT method identifies split points in this period, the right-most point is used to split the series; otherwise, the $l$ remains the original split point. Hence, we obtain subspaces varying in length between $l$ and $l+t$. This is illustrated in Figure~\ref{fig:trace_segmentation_ex}. 
Finally, we re-interpolate the obtained segments to have the same length $l+t$~\cite{dtwtutorial}, which is required to be able to pre-compute the Keogh envelopes.


\section{Data mining applications}
Similarity comparisons between pairs of series are a core subroutine in most time series data mining approaches. In this section, we discuss how PQDTW can be incorporated in 1-NN and hierarchical clustering.

\subsection{NN search with PQ approximates}
A large body of empirical research has shown that NN-DTW is the method of choice for most time series classification problems~\cite{ding2008,tan2019elastic,paparrizos2020}. However, being a lazy learner, the main drawback is its time and space complexity. The entire training set has to be stored and the classification time is a function of the size of this training set. Therefore, NN-DTW still has severe tractability issues in some applications. This is especially true for resource-constrained devices such as wearables.

NN search with PQ is both fast (only the query has to be encoded online and only $M$ additions are required per distance calculation) and reduces significantly the memory requirements for storing the training data. It proceeds as follows: Given a query time series $y \in \mathbb{R}^D$ and a database of $N$ time series $X = \{x_n\}_{n=1}^N$ where each $x_n \in \mathbb{R}^D$. At training time, a PQ is first trained on (a subset of) the database time series, which are subsequently encoded (section~\ref{sec:encoding}) using the trained PQ. At prediction time, the approximate distance is computed between the query vector $y$ and each encoded database time series. In most cases, one should use the asymmetric version, which obtains a lower distance distortion. When the training set is large, the $O(D \times K)$ DTW computations required to compute the asymmetric distance look-up table is relatively small. The only exceptions are queries in small databases (albeit, techniques other than PQ are more appropriate in such cases) or applications with many time-critical queries.

The linear scan with PQDTW is fast compared to the state-of-the-art NN-DTW methods~\cite{rakthanmanon2012lbcascade}, but still slow for a large number of $N$. To handle million-scale search, a search system with inverted indexing was developed in the original PQ paper~\cite{jegou2010pq}.

\subsection{Clustering with PQ approximates}
\label{sec:clustering}
While all of the conventional clustering approaches 
rely on similarity comparisons in which DTW can be substituted by PQDTW, we focus on the hierarchical algorithms in this paper. These have great visualization power in time series clustering
and do not require the number of clusters as an additional parameter. However, at the same time, hierarchical clustering does not scale to large datasets, because it requires the computation of the full pairwise distance matrix. Therefore, lower-bound pruning cannot be applied.

For constructing a pairwise distance matrix, asymmetric distance computation is an expensive operation since it involves the computation of the full DTW distance matrix between the subspaces of each time series in the dataset and the codebook (i.e., $N \times K \times M$ DTW computations). This is only acceptable if the number of time series is a lot bigger than the number of centroids in the codebooks.

Using symmetric distance computation, two similar time series have a high likelihood to be mapped to the same centroids, resulting in an approximate distance of zero. While the resulting errors on the estimated distance are small, this might be problematic in clustering applications where the ranking of distances is important. This is solved by partially replacing the estimated distance when subspaces are encoded to the same code. As an efficient and elegant replacement value, we propose the Keogh lower bound. Given a sub-quantizer $q_m$ and two subspaces $x^m$ and $y^m$, the distance value would be $max(lb(x^m,q_m(y^m)),lb(q_m(x^m),y^m))$. This bound is guaranteed to be between 0 and the exact distance.

%% file: chapters/04-experiments.tex
\begin{figure*}[t]
     \centering
     \begin{subfigure}[t]{0.32\textwidth}
         \centering
         \includegraphics[width=\textwidth]{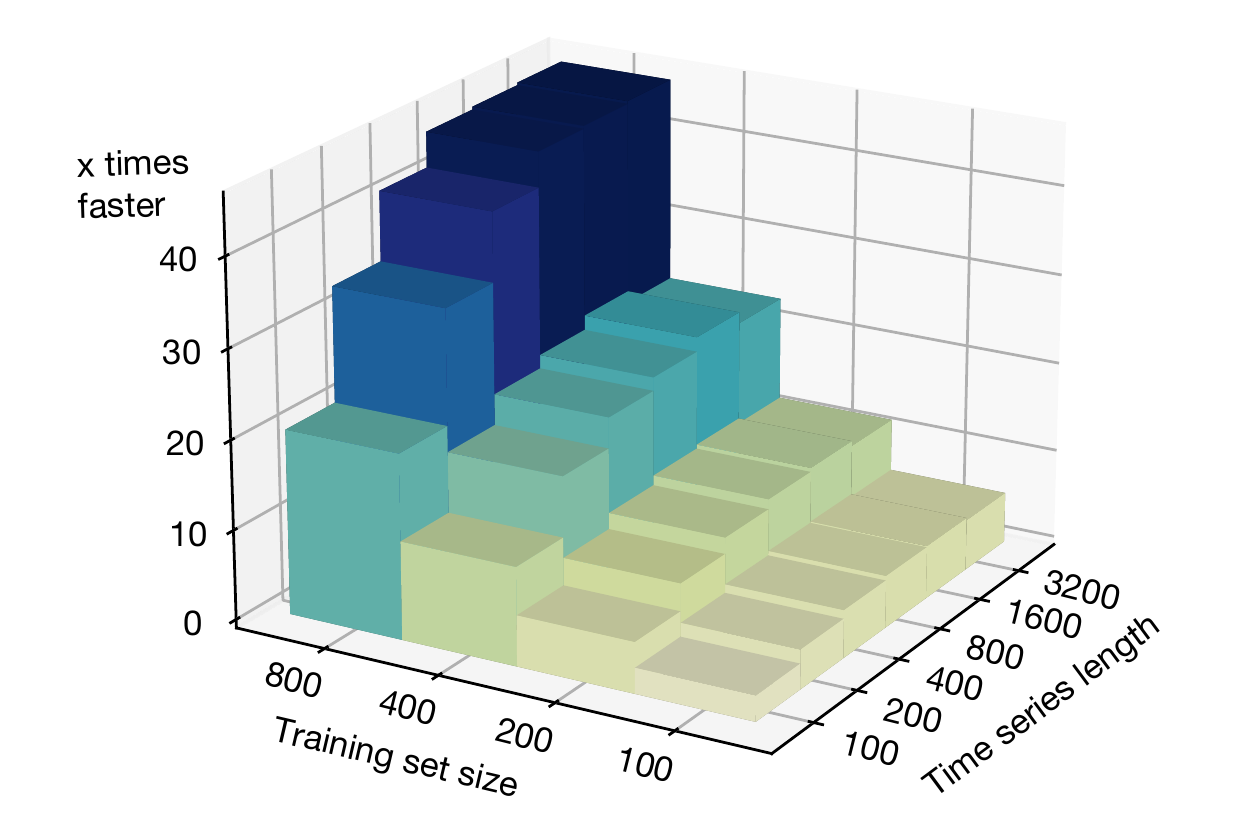}
         \caption{Comparison of the runtime of PQDTW (no pre-alignment) with DTW}
         \label{fig:pqdtw_time_complexity_a}
     \end{subfigure}
     \hfill
     \begin{subfigure}[t]{0.32\textwidth}
         \centering
         \includegraphics[width=\textwidth]{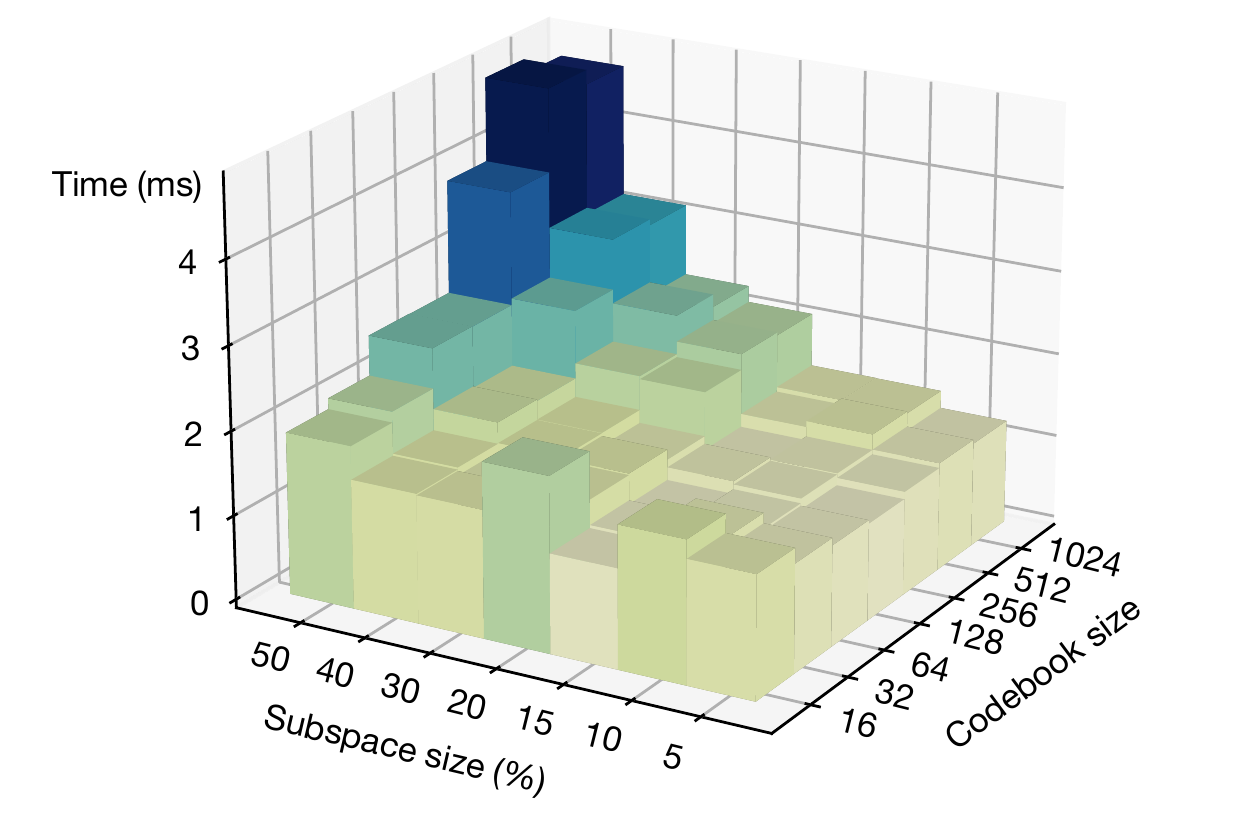}
         \caption{Effect of the subspace and codebook size on the runtime of PQDTW}
         \label{fig:pqdtw_time_complexity_b}
     \end{subfigure}
     \hfill
     \begin{subfigure}[t]{0.32\textwidth}
         \centering
         \includegraphics[width=\textwidth]{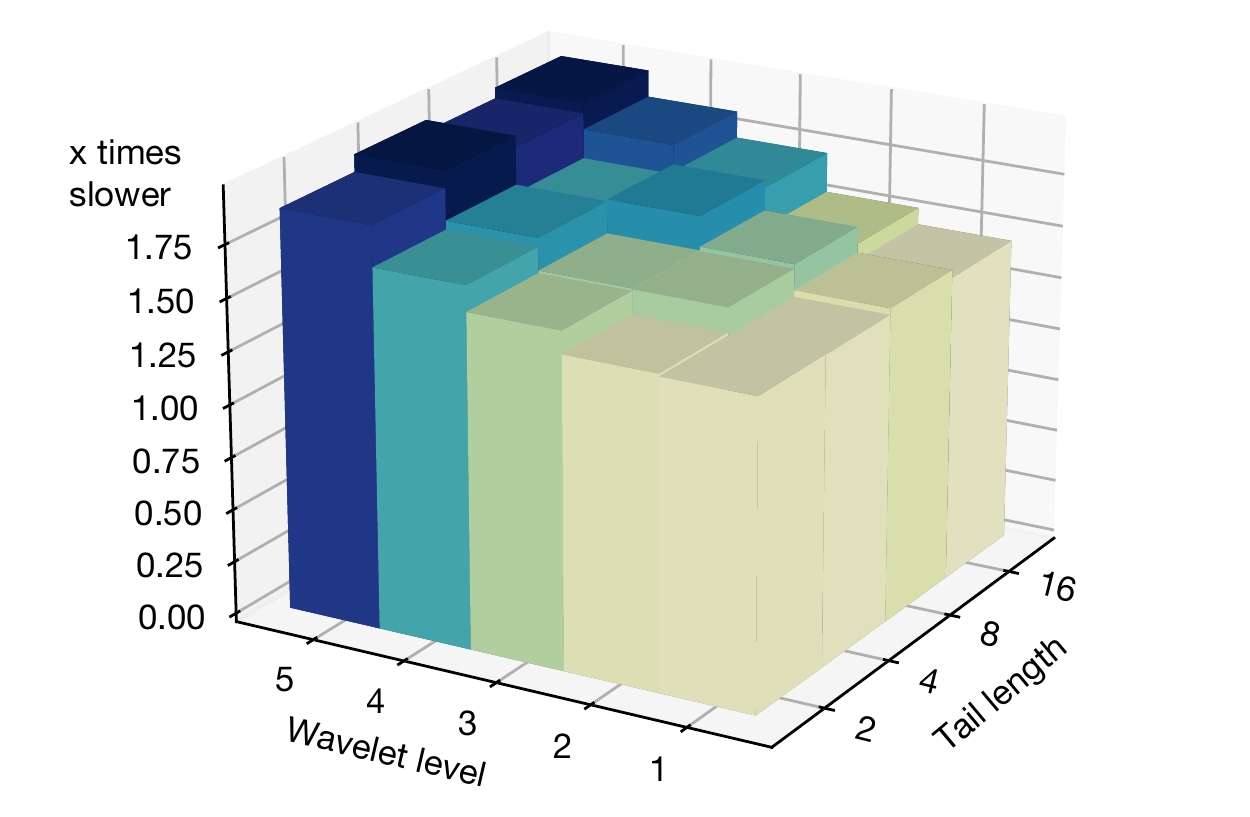}
         \caption{Effect of pre-alignment on the runtime of PQDTW}
         \label{fig:pqdtw_time_complexity_c}
     \end{subfigure}
        \caption{Empirical evaluation of the time complexity of PQDTW on a random walks dataset.}
        \label{fig:pqdtw_time_complexity}
\end{figure*}

\section{Experimental Settings}
This section describes the experimental settings for the evaluation of PQDTW.

\textbf{Platform} We ran all experiments on a set of identical computing servers\footnote{Intel Core i7-2600 CPU @ 3.40GHz; 15Gb of memory; Ubuntu GNU/Linux 18.04} using a single core per run. In order to reduce the variance in runtime caused by other processes outside our control and the variance caused by the random selection of centroids in the DBA k-means step of the PQDTW encoding step, we executed each method five times with different seeds and report the mean accuracy and median run time.

\textbf{Baselines} We compare PQDTW against the most common and state-of-the-art distance measures for time series: Euclidean distance (ED), dynamic time warping (DTW)~\cite{sakoe1978}, window-constrained dynamic time warping (cDTW)~\cite{sakoe1978}, and shape-based distance (SBD)~\cite{paparrizos2015kshape}. For DTW we use the PrunedDTW~\cite{silva2016pruneddtw} technique to prune unpromising alignments. For constrained DTW, we consider window sizes of 5 and 10\%, as well as the window size which leads to the minimal 1NN classification error on the training set. We denote the window size with a suffix (e.g., cDTW10) and using cDTWX for the optimal one. SBD is a state-of-the-art shape-based distance measure, achieving similar results to cDTW and DTW while being orders of magnitude faster. In addition, we compare against SAX~\cite{lin2003:sax}, which is perhaps the most studied symbolic representation for time series. We use an alphabet size $\alpha = 4$, and segments of length $l = 0.2 * L$ (where L is length of the time series)~\cite{nguyen2017}. Finally, we compare against standard PQ using the Euclidean distance (i.e., a version of PQDTW without pre-alignment that uses ED instead of DTW), denoted PQ\textsubscript{ED}.

\textbf{Implementation} We implemented PQDTW, ED, DTW, and SBD under the same framework, in C(ython)~\cite{Meert_DTAIDistance_2020}, for a consistent evaluation in terms of both accuracy and efficiency. For repeatability purposes, we make all source code available.\footnote{ \url{https://github.com/probberechts/PQDTW}}
For SAX, we use the Cython implementation available in {\tt tslearn}.\footnote{We use {\it tslearn v0.5.0.5}. See \url{tslearn.readthedocs.io}.}

\textbf{Parameter settings} A disadvantage of the PQDTW approach is that we have many hyper-parameters to tune. We use a default codebook size of 256 (or all time series in the training set if there are less examples) and symmetric distance computation. To determine the optimal subspace size, wavelet level, tail and quantization window, we use the Tree-structured Parzen Estimator algorithm~\cite{optuna2019} which we ran for 12 hours on each dataset. This hyper-parameter tuning is a one-time effort. We use 5-fold cross validation on the training set with a test set of 25\% and evaluate the 1NN classification error. This results in multiple Pareto optimal solutions with respect to runtime and accuracy. We report the results for the most accurate solution on the training set.

\textbf{Statistical analysis} We analyze the results of every pairwise comparison of algorithms over multiple datasets using the Friedman test followed by the post-hoc Nemenyi test. We report statistical significant results with a 95\% confidence level.

\section{Experimental Results}

\begin{table*}[h!]
\caption{Comparison of PQDTW against other distance measures for 1NN and hierarchical complete linkage clustering. The column ``Mean difference'' contains the mean  and standard deviation of the relative difference in classification error (1NN) and rand index (clustering) between PQDTW and the corresponding measure, whereas ``Speedup'' indicates the factor by which PQDTW speeds up the runtime.}
\label{table:comp}
\def\sym#1{\ifmmode^{#1}\else\(^{#1}\)\fi}
\sisetup{separate-uncertainty,table-space-text-post = \sym{*}}
\begin{tabular*}{\textwidth}{
@{\extracolsep{\fill}}l
S[table-format = -1.3(2),table-align-text-post=false]
r
S[table-format = -1.3(2),table-align-text-post=false]
r
}
\toprule
 & \multicolumn{2}{c}{\bf 1NN} & \multicolumn{2}{c}{\bf Clustering} \\ 
 & \thead{Mean Error difference} & \thead{Speedup} & \thead{Mean ARI difference} & \thead{Speedup} \\ 
\midrule
ED              & 0.017(066) & x14.00 &  0.013(099) &  x2.64  \\  \midrule
DTW             & - 0.014(064)\sym{*} & x25.01 & 0.004(122) & x225.20 \\
cDTW5           & - 0.036(044)\sym{*} & x12.91 & 0.008(097) & x32.83   \\
cDTW10          & - 0.029(052)\sym{*} & x15.81  & -0.002(110) & x59.01   \\
cDTWX           & - 0.037(050)\sym{*} & x14.15 & -0.003(106) &  x50.45  \\ 
SBD             & - 0.021(056)\sym{*} & x6.45 & -0.011(105) & x47.18  \\ \midrule
SAX             &  0.293(199)\sym{+} & x190.63 & 0.043(140) &  x884.77  \\ 
PQ\textsubscript{ED} & 0.038(071)\sym{+} & x0.83 & -0.006(057) & x0.75  \\ 
\bottomrule
\multicolumn{5}{l}{
    \footnotesize 
    \sym{*} PQDTW performs worse ($p<0.05$);
    \sym{+} PQDTW performs better ($p<0.05$)
}
\end{tabular*}
\end{table*}

The goal of this evaluation is to demonstrate the efficiency and accuracy of PQDTW in classification and clustering applications. This evaluation will first demonstrate the empirical time complexity of PQDTW in comparison to DTW and evaluate the effect of parameter settings on a synthetically generated random walk dataset. Second, we benchmark PQDTW for time series classification and clustering on 48 UCR datasets.\footnote{Only the datasets available since 2018~\cite{UCRArchive2018} were used to keep the runtime of the experiments manageable, while achieving a maximal overlap with existing research.}

\subsection{Empirical time complexity}
We begin with an evaluation of the empirical time complexity on a random walk dataset. Although these random walks are not ideal to evaluate the PQDTW algorithm due to lack of common structures that can be aligned, they allow us to do reproducible experiments on a large set of time series collections of varying sizes and time series lengths. The results show a significant speedup of PQDTW (subspace size = $20\%$, no pre-alignment) over DTW, which improves relatively for longer time series (Figure~\ref{fig:pqdtw_time_complexity_a}). For computing the pairwise distance matrix of 100 time series, PQDTW is between 2.9 times (length 100) and 5.6 times faster (length 3200). Interestingly, due to lower bound pruning, the average computation time of PQDTW per pair of time series decreases a lot if the number of time series grows. Therefore, for a collection of 800 time series of length 3200, PQDTW is already 45.8 times faster.

The parameters that affect the speed of PQDTW most are the subspace size and codebook size (Figure~\ref{fig:pqdtw_time_complexity_b}). In accordance with the theoretical time complexity $O(K\times D^2/M)$, the runtime increases linearly when less subpaces or a larger codebook is used. 

Finally, the pre-alignment step has a minor effect on the runtime (Figure~\ref{fig:pqdtw_time_complexity_c}), which is mainly determined by the level of the wavelet decomposition. Increasing the tail length does not have a significant effect. 


\subsection{1NN classification}
Table~\ref{table:comp} reports the classification error and runtime of PQDTW against the state-of-the-art distance measures. For DTW and cDTW, we use the Keogh lower bound for early stopping. The statistical test suggests that there is no significant difference between PQDTW and ED. PQDTW performs at least as well in 23 datasets. All other distance measures that operate on the raw data outperform PQDTW with statistical significance. However, Figure~\ref{fig:comp_pqdtw_dtw} shows that the difference in accuracy between PQDTW and cDTWX (i.e., the best performing measure) is small in all cases, while PQDTW is 14x faster on average. Additionally, PQDTW compresses the training data by a factor varying between 26.2 and 2622.4, depending on the dataset and PQDTW's parameter settings. From this experiment, we can conclude that (1) PQDTW is competitive with ED, but is much faster and requires far less space; and that (2) PQDTW outperforms SAX and PQ\textsubscript{ED}, the baseline dimensionality reduction techniques based on ED.

\subsection{Hierarchical clustering}
\begin{figure}
    \centering
    \begin{subfigure}[t]{0.4\columnwidth}
        \centering
        \includegraphics[width=\textwidth,trim={2cm 0 3cm 0},clip]{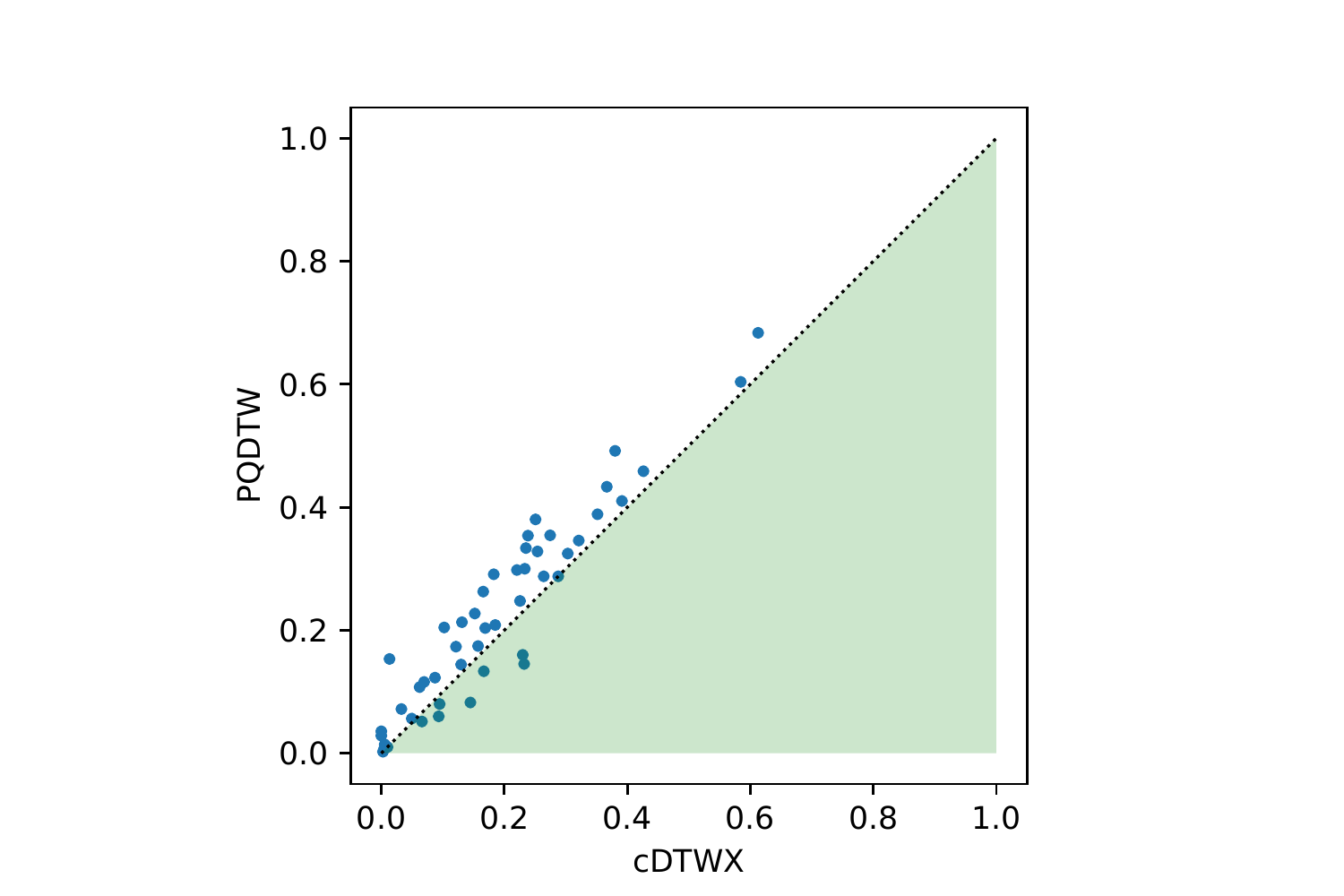}
        \caption{1NN}
        \label{fig:comp_nn_pqdtw_dtw}
    \end{subfigure}%
    ~ 
    \begin{subfigure}[t]{0.4\columnwidth}
        \centering
        \includegraphics[width=\textwidth,trim={2cm 0 3cm 0},clip]{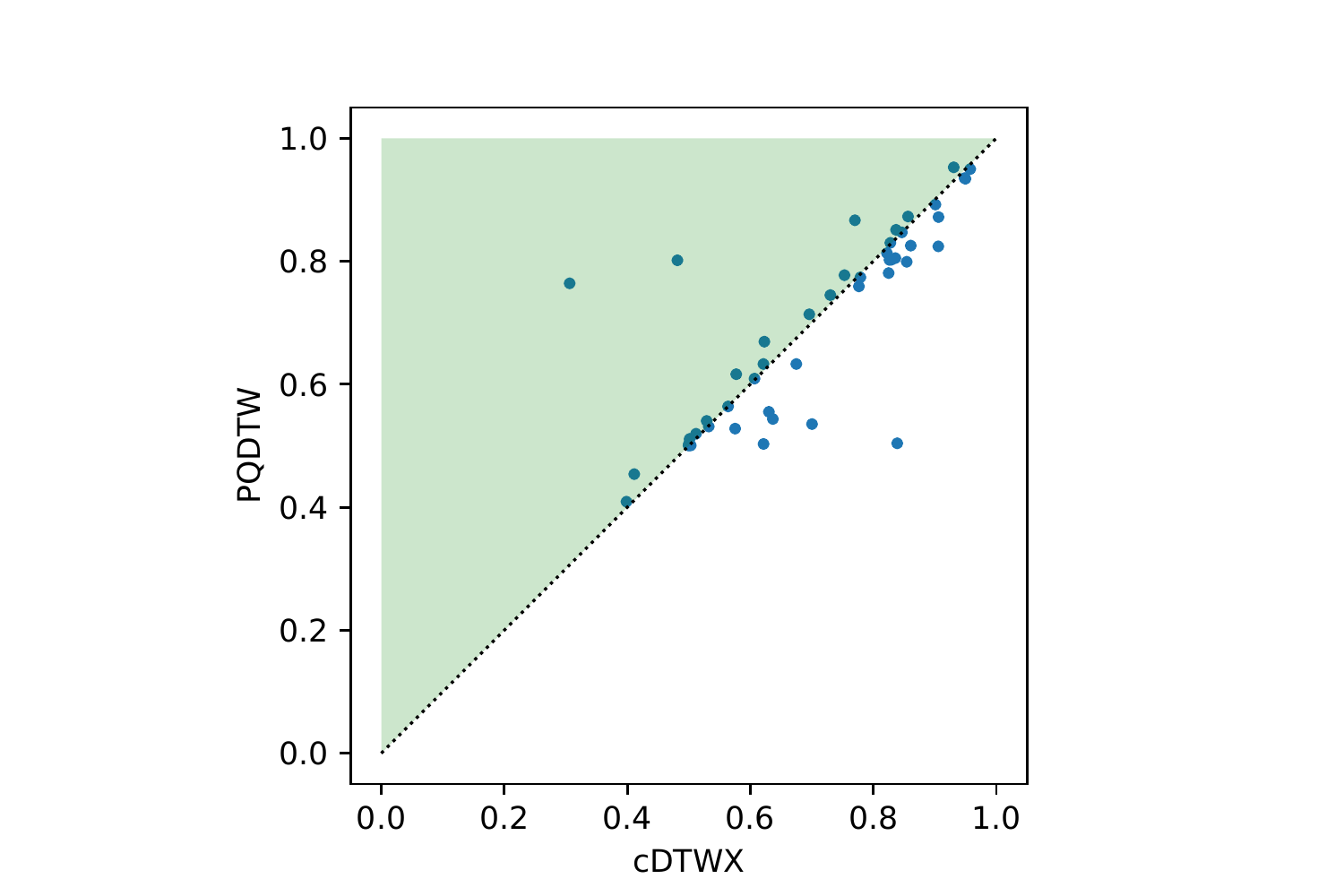}
        \caption{Clustering}
        \label{fig:comp_clust_pqdtw_dtw}
    \end{subfigure}
    \caption{Comparison of (a) the 1NN classification error and (b) the rand index for hierarchical complete linkage clustering with PQDTW and cDTWX over 48 UCR datasets. Circles in the green area indicate datasets for which PQDTW performs better than cDTWx.}
    \label{fig:comp_pqdtw_dtw}
\end{figure}

We use agglomerative hierarchical clustering with single, average, and complete linkage criteria. To evaluate the obtained clustering, we compute a threshold that cuts the produced dendrogram at the minimum height such that $k$ clusters are formed, with $k$ corresponding to the number of classes in the dataset. Subsequently, we compute the Rand Index (RI)~\cite{rand1971} over the test set using the class labels as the ground truth clustering. The major difference in performance among hierarchical methods is the linkage criterion and not the distance measure~\cite{paparrizos2015kshape}. Since the complete linkage criterion gave the best results, we only report these in Table~\ref{table:comp}. There are no significant differences among all distance measures that we evaluated. Figure~\ref{fig:comp_clust_pqdtw_dtw} shows that the differences in RI between PQDTW and cDTWX are indeed small. Since the full distance matrix has to be computed to obtain a hierarchical clustering and lower bound pruning cannot be applied, the gain in performance is larger compared to 1NN. Our approach is one order of magnitude faster than cDTW and SBD, and two orders of magnitude faster than DTW.

%% file: chapters/07-conclusion.tex
\section{Conclusions}
This work presented PQDTW, a generalization of the product quantization algorithm for Euclidean distance to DTW. By exploiting prior knowledge about the data through quantization and compensating for subtle variations along the time axis through DTW, PQDTW learns a tighter approximation of the original time series than other piecewise approximation schemes proposed earlier. 
Overall, the results suggest that PQDTW is a strong candidate for time series data analysis applications in online settings and in situations where computation time and storage demands are an issue.